\documentclass[10pt, conference, a4paper]{IEEEtran}
\IEEEoverridecommandlockouts
\usepackage{cite}
\usepackage{amsmath,amssymb,amsfonts}
\usepackage{bm}
\usepackage{mathtools}
\usepackage{nicefrac}
\usepackage{algorithmic}
\usepackage{graphicx}
\usepackage{textcomp}
\usepackage{xcolor}
\def\BibTeX{{\rm B\kern-.05em{\sc i\kern-.025em b}\kern-.08em
    T\kern-.1667em\lower.7ex\hbox{E}\kern-.125emX}}
\usepackage{lipsum}
\usepackage{pgfplots}

\usetikzlibrary{positioning,calc}
\usetikzlibrary{arrows.meta}

\pgfplotsset{compat=1.16}
\pgfplotsset{table/search path={data}}

\usepgfplotslibrary{groupplots}
\usepgfplotslibrary{polar}

\graphicspath{{images/}}

\definecolor{DLRBlack}{gray}{0}
\definecolor{DLRGrey}{gray}{0.420} % 107/255
\colorlet{DLRGray}{DLRGrey}
\definecolor{DLRWhite}{gray}{1}

\colorlet{DLRDarkerGrey}{DLRGrey}
\definecolor{DLRDarkGrey}{gray}{0.537} % 137/255
\definecolor{DLRMediumGrey}{gray}{0.702} % 179/255
\definecolor{DLRLightGrey}{gray}{0.820} % 209/255
\definecolor{DLRLighterGrey}{gray}{0.929} % 237/255

\colorlet{DLRDarkerGray}{DLRDarkerGrey}
\colorlet{DLRDarkGray}{DLRDarkGrey}
\colorlet{DLRMediumGray}{DLRMediumGrey}
\colorlet{DLRLightGray}{DLRLightGrey}
\colorlet{DLRLighterGray}{DLRLighterGrey}

\definecolor{DLRDarkerBlue}{RGB}{0, 106, 144}
\definecolor{DLRDarkBlue}{RGB}{0, 156, 208}
\definecolor{DLRBlue}{RGB}{33, 187, 223}
\colorlet{DLRMediumBlue}{DLRBlue}
\definecolor{DLRLightBlue}{RGB}{149, 212, 238}
\definecolor{DLRLighterBlue}{RGB}{201, 232, 251}

\definecolor{DLRDarkerGreen}{RGB}{115, 163, 63}
\definecolor{DLRDarkGreen}{RGB}{158, 193, 76}
\definecolor{DLRGreen}{RGB}{199, 214, 84}
\colorlet{DLRMediumGreen}{DLRGreen}
\definecolor{DLRLightGreen}{RGB}{215, 223, 116}
\definecolor{DLRLighterGreen}{RGB}{228, 234, 173}

\definecolor{DLRDarkerYellow}{RGB}{224, 177, 57}
\definecolor{DLRDarkYellow}{RGB}{254, 206, 73}
\definecolor{DLRYellow}{RGB}{255, 223, 73}
\colorlet{DLRMediumYellow}{DLRYellow}
\definecolor{DLRLightYellow}{RGB}{255, 234, 117}
\definecolor{DLRLighterYellow}{RGB}{255, 248, 189}

\newcommand{\mW}{\text{mW}}

\title{%
Seeing the World through an Antenna's Eye:   
Reception Quality Visualization Using Incomplete Technical Signal Information
}

\author{
\IEEEauthorblockN{Leif Bergerhoff}
\IEEEauthorblockA{%
\textit{German Aerospace Center (DLR)}\\
\textit{German Remote Sensing Data Center (DFD)}\\
\textit{National Ground Segment (NBS)}\\
Neustrelitz, Germany\\
Leif.Bergerhoff@dlr.de}
}

\begin{document}

\maketitle

%-------------------------------------------------------------------------------

\begin{abstract}
We come up with a novel application for image analysis methods in the context 
of direction dependent signal characteristics.
For this purpose, we describe an inpainting approach adding benefit to 
technical signal information which are typically only used for monitoring and 
control purposes in ground station operations.
Recalling the theoretical properties of the employed inpainting technique and 
appropriate modeling allow us to demonstrate the usefulness of our approach 
for satellite data reception quality assessment.
In our application, we show the advantages of inpainting products over raw data 
as well as the rich potential of the visualization of technical signal 
information. 
\end{abstract}

\begin{IEEEkeywords}
signal information, satellite data reception, inpainting, homogeneous 
diffusion, partial differential equations, data analysis, modeling, application
\end{IEEEkeywords}

%-------------------------------------------------------------------------------

\section{Introduction}
A fundamental part of daily operations at a ground station is the surveillance and 
evaluation of the data exchange with satellites.
This requires good knowledge of the status and health of all equipment and 
services which are part of the reception and transmission chains. 
Due to the tremendous system complexity, the resulting high dimensionality of the 
available 
raw information, and the large number of different satellite contacts, a reliable 
manual quality and error analysis is not feasible if impossible.
As a real-time data center for remote sensing and small scientific satellites,
the ground station Neustrelitz was involved in about 17 000 operational satellite 
contacts in more than 26 missions in 2023. 
This includes earth observation satellites like EnMAP \cite{St23}, TerraSAR-X 
\cite{SMSL16}, and Landsat 9 \cite{MBDEHJKKKMMMPPW19} as well as DSCOVR 
\cite{BB12} in a space weather context.
Currently, operational communication with satellites results in about 19~GiB of 
device and signal specific measurements every year.
These numbers underline the necessity of advanced signal analysis techniques 
for the processing of technical signal information in a real-world scenario.
Additionally, the available data is often only available in parts and not for a 
complete domain which represents another challange.
Signal level information for antenna pointing directions serve as a good 
example here.
They are not available for the 
whole range of possible directions but only for those of past satellite orbits.

Within the image analysis community, so-called inpainting techniques are a common 
tool to complete missing information. The technique of homogeneous diffusion 
inpainting suits well to restore image data from edges \cite{Ca88,El99,HM89,RZ86} 
or based on selected data points \cite{JSGA86,KLDJPFR05,LNG03}.
Another application is the field of image compression \cite{Ma15}.
In a recent publication \cite{PSAW23}, Peter et al. relate traditional homogeneous 
diffusion inpainting and neural networks.

\subsection{Our Contributions}
We present a novel approach which links technical signal information from  
ground station operations with well established tools from image analysis.
The presented method employs homogeneous diffusion inpainting to create a 
two-dimensional visualization of directional dependent measurements from 
incomplete data.
In our application, we demonstrate the usefulness of our approach for reception 
quality assessment. As input data, we use signal level information from 
operational S-band \cite{IEEE19} data reception as well as the corresponding 
antenna pointing directions. We visualize the inpainted result and show its 
potential for exploration and assessment of technical signal information.
To the best of our knowledge, this is the first attempt to combine ground 
station metadata and inpainting techniques.
We see our contribution as a starting point for a comprehensive analysis of the 
technical signal information gained in operations with the aim of ground 
station development.

\subsection{Structure of the Paper}
In Section~\ref{sec:signal_info}, we give a brief introduction to the technical 
signal information which motivate the development of our method and serve as a 
data basis for our application.
Section~\ref{sec:theory} contains the mathematical background of homogeneous 
diffusion inpainting for two-dimensional direction dependent data.
Section~\ref{sec:app} presents a real-world application and gives results of 
inpainted signal level information for antenna pointing directions.
We conclude with a summary and outlook in Section~\ref{sec:concl}.

%-------------------------------------------------------------------------------

\section{Technical Signal Information}
\label{sec:signal_info}
Typically, communications with earth observation satelites at the ground 
station Neustrelitz last slightly more than 12 minutes. During this time, the 
related hardware and software provide status information about the currently 
received signal and the extracted data.
This includes
\begin{itemize}
\item antenna systems \textit{(e.g. antenna pointing information)}, 
\item modems \textit{(e.g. signal level information)}, and
\item front end processors \textit{(e.g. extracted telemetry data information)}.
\end{itemize}
Subsequently, we refer to this data as technical signal information.
Usually, information is available as time-series data with a sampling interval 
of one second. I.e. a 12 minute satellite contact results in 720 measurements 
for each device involved in the communication process. In our case, one 
measurement can contain values for up to 80 parameters in parallel.
\begin{figure}
\centering
% Landsat-9 10439
% support: 2023-09-14 10:51:43.0 - 2023-09-14 11:03:01.0
%
%
\begin{tikzpicture}
\begin{polaraxis}[%
    width=.75\columnwidth,
    height=.75\columnwidth,
    rotate=-90,
    x dir=reverse,
    xmin=0,
    xmax=360,
    ymin=0,
    ymax=100,
    xtick={0,45,90,135,180,225,270,315},
    xticklabels={N,\empty,E,\empty,S,\empty,W,\empty},
    ytick={0,30,60,90},
    yticklabels={$90^\circ$,$60^\circ$,$30^\circ$,$0^\circ$},
    ytick align=center,
    yticklabel style={%
        anchor=north,
        xshift=3*\pgfkeysvalueof{/pgfplots/major tick length},
        yshift=2.5*\pgfkeysvalueof{/pgfplots/major tick length}},
    axis line style={draw=none},
    grid style={line width=.5pt, draw=DLRMediumGrey},
    every axis title/.style={at={(-0.25,0.75)}, anchor=north},
    /tikz/font=\tiny]
\addplot+[DLRDarkerBlue, mark=none, thick] table [x=az, y=el_inv]
{landsat9_10439_ant.tab};
\end{polaraxis}
\end{tikzpicture}
\vskip 2mm
\begin{tikzpicture}
\begin{axis}[%
    width=\columnwidth,
    height=.45\columnwidth,
    xmin=39103, % 10:51:43
    xmax=39781, % 11:04 11:03:01
    restrict x to domain=39103:39781,
    scaled x ticks=false,
    xtick={39120,39180,39240,39300,39360,39420,39480,39540,39600,
        39660,39720,39780},
    xticklabels={{10:52},{10:53},{10:54},{10:55},{10:56},{10:57},{10:58},
        {10:59},{11:00},{11:01},{11:02},{11:03}},
    x tick label style={rotate=90},
    xlabel={UTC [HH:MM]},
    /tikz/font=\tiny,
    ylabel={Signal Level [dBm]},
    grid=both,
    minor tick num=1]
\addplot+[DLRDarkerBlue, mark=none, thick]
table [x={time}, y={if_lvl}]
{landsat9_10439_crt.tab};
\end{axis}
\end{tikzpicture}
\caption{An excerpt of technical signal information from a Landsat 9 satellite 
    contact in September 2023. \textbf{Top:}~Antenna pointing. 
    \textbf{Bottom:}~Measured Signal level.}
\label{fig:example_ls9}
\end{figure}
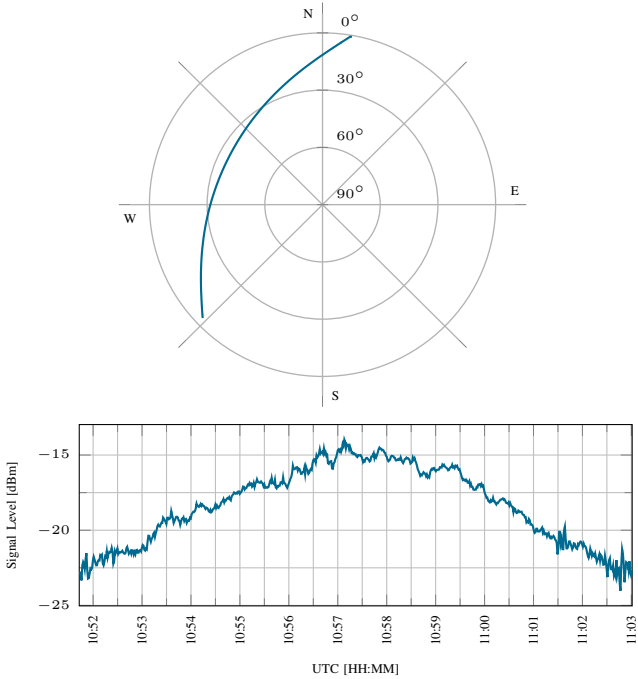
In Fig.~\ref{fig:example_ls9}, we provide 
an example for technical signal information which we store in our database. On 
top, one can see the antenna pointing in terms of the elevation $\theta$ (0 to 
90 degrees) and azimuth $\varphi$ (0 to 360 degrees) angle. The corresponding 
signal level over time is illustrated below.
The signal levels are given in decibel-milliwatts (dBm), i.e. given a power $P$ 
in milliwatts (mW), the corresponding power level $x$ in dBm is given by:
\begin{equation} \label{eq:dbm}
    x = 10 \cdot \log_{10} \frac{P}{1~\mW}\,.
\end{equation}
Usually, technical signal information serve for monitoring and control purposes 
of the ground station during satellite data reception and transmission. 
Subsequently, we show their additional benefit for reception quality assessment.

%-------------------------------------------------------------------------------

\section{Modeling / Theory}
\label{sec:theory}

\subsection{Problem Description}
Let $\bm{x} = (\theta, \varphi)^\top$ denote an antenna pointing direction, 
whereas $\theta \in [0, \nicefrac{\pi}{2}]$ represents the elevation angle and
$\varphi \in [0,2\pi]$ represents the azimuth angle. Based on this, we define 
the  
two-dimensional domain spanned by azimuth and elevation as
\begin{equation}
\Omega :=  \{ \bm{x} \, \big| \, \forall \,
\theta \in [0, \tfrac{\pi}{2}], \, \varphi \in [0, 2 \pi] \}.
\end{equation}
For the set $\Omega_K \subset \Omega$ of known data, we have signal level 
information
\begin{equation}
f(\bm{x}): \quad \Omega_K \to \mathbb{R},
\end{equation}
available. $\Omega_K$ is also called \textit{inpainting data}.
Our aim is to fill in the missing signal level information in the so-called 
\textit{inpainting domain} $\Omega \setminus \Omega_K$ based on $f$. This 
process of 
filling in is called \textit{inpainting}.

\subsection{Homogeneous Diffusion Inpainting}
To estimate the missing data, we make use of a well-known and elegant technique 
called homogeneous diffusion inpainting.
Within this context, we describe the desired signal level at time $t$ as
$u := u(\bm{x}, t)$ with
\begin{equation}
u(\bm{x}, t): \quad \Omega \times [0, \infty) \to \mathbb{R} \,.
\end{equation}
The time $t$ serves as a model parameter and we focus on the scenario $t \to 
\infty$.
Using the known data $f$ as Dirichlet boundary conditions, we estimate $u$ by 
solving the Laplace 
equation on the set of \textit{unknown} data $\Omega \setminus \Omega_K$:
\begin{align}
u(\bm{x}) & = f(\bm{x}) & & \text{for} \,\, \bm{x} \in \Omega_K,\\
\Delta u (\bm{x}) &= 0 & & \text{for} \,\, \bm{x} \in \Omega \setminus \Omega_K,
\end{align}
Referring to the real-world antenna pointing directions, we make use of mixed 
boundary conditions that model the periodicity of the azimuth domain and the point 
symmetry w.r.t. the point $(\nicefrac{\pi}{2},\varphi)$. Furthermore, we take into 
account homogeneous Neumann boundary conditions for $\theta = 0$.
Consequently, we have:
\begin{align}
\label{eq:bc0}
\partial_\theta u \big|_{\theta = 0} & = 0,\\[.5ex]
\label{eq:bc1}
u(\tfrac{\pi}{2} + \Delta \theta, \varphi, t) & =
u(\tfrac{\pi}{2} - \Delta \theta, \varphi + \pi, t),\\[.5ex]
\label{eq:bc_phi}
u (\theta, 2 \pi, t) & = u (\theta, 0, t) \qquad \forall \, \theta \in [0, 
\frac{\pi}{2}] .
\end{align}
Keeping this in mind, we can write our inpainting problem as
\begin{equation} \label{eq:inp_problem}
    c(\bm{x}) (u(\bm{x}) - f(\bm{x})) -
    (1 - c(\bm{x})) \Delta u (\bm{x}) = 0\,.
\end{equation}
The confidence function $c(\bm{x})$ -- also referred to as \textit{inpainting 
mask} 
-- indicates whether a point is known or not:
\begin{equation} \label{eq:confidence}
c(\bm{x}) =
\begin{cases}
    1 & \text{for} \quad \bm{x} \in \Omega_K,\\
    0 & \text{for} \quad \bm{x} \in \Omega \setminus \Omega_K.
\end{cases}
\end{equation}
The Laplacian of $u$ is given by
\begin{equation}
\Delta u = \partial_{\theta \theta} u + \partial_{\varphi \varphi} u \,.
\end{equation}

\subsection{Discrete Theory}
Our goal is to solve the inpainting problem on the discrete 
elevation-azimuth domain which we assume to be a regular two-dimensional grid. 
In total, we make use of $N$ different pointing directions, i.e. 
$(\theta,\varphi)_i$ for $i \in \{1,...,N\}$.
Storing the two-dimensional information row-wise, we represent the discrete 
version of the known data $f$ as a one-dimensional vector $\bm{f}$ and the desired 
signal $u$ by means of the one-dimensional vector $\bm{u}$.
Furthermore, we denote the $i$-th element in $\theta$ 
and the $j$-th element in $\varphi$ direction by $f_{i,j}$ and $u_{i,j}$. 
Using a binary pixel mask $\bm{c}$ as the discrete counterpart of 
\eqref{eq:confidence}, 
we express whether signal information is known or not, i.e. $c_i =0$ for unknown 
pixels and $c_i = 1$ for known pixels.
Consequently, the discrete version of \eqref{eq:inp_problem} reads
\begin{equation} \label{eq:disc_inpaint}
\mathbf{C} (\bm{u} - \bm{f}) - (\mathbf{I} - \mathbf{C}) \mathbf{A} \bm{u} = 0\, ,
\end{equation}
where $\mathbf{I}$ is the identity matrix and $\mathbf{C} := \text{diag}(\bm{c})$ 
is a diagonal matrix containing the elements of $\bm{c}$ on its diagonal.
Considering the boundary conditions \eqref{eq:bc0}--\eqref{eq:bc_phi}, the 
symmetric $N \times N$ matrix $\mathbf{A}$ implements the discrete Laplace 
operator $\Delta_{i,j}$ by means of central finite differences and the grid 
sizes 
$h_\theta$ and $h_\varphi$, where
\begin{align}
\partial_{\theta \theta} u_{i,j} & \approx
\frac{u_{i+1,j} - 2 u_{i,j} + u_{i-1,j}}{h_\theta^2}\, ,\\
\partial_{\varphi \varphi} u_{i,j} & \approx
\frac{u_{i,j+1} - 2 u_{i,j} + u_{i,j-1}}{h_\varphi^2} \, .
\end{align}
A minimal example which illustrates the boundary conditions for $\bm{u}$ can be 
found in 
Table~\ref{tab:bcond}.
\begin{table}
\caption{A minimal example illustrating the boundary conditions in our discrete 
setup.}
\label{tab:bcond}
\centering
\begin{tabular}{c||c|c|c|c||c}
& $u_{4,2}$  & $u_{4,3}$  & $u_{4,0}$  & $u_{4,1}$  & \\
\hline\hline
$u_{4,3}$ & ${u_{4,0}}$  & ${u_{4,1}}$  & ${u_{4,2}}$  & ${u_{4,3}}$ & 
$u_{4,0}$\\\hline
$u_{3,3}$ & ${u_{3,0}}$  & ${u_{3,1}}$  & ${u_{3,2}}$  & ${u_{3,3}}$ & 
$u_{3,0}$\\\hline
$u_{2,3}$ & ${u_{2,0}}$  & ${u_{2,1}}$ & ${u_{2,2}}$ & ${u_{2,3}}$ & 
$u_{2,0}$\\\hline
$u_{1,3}$ & ${u_{1,0}}$ & ${u_{1,1}}$ & ${u_{1,2}}$ & ${u_{1,3}}$ & 
$u_{1,0}$\\\hline
$u_{0,3}$ & ${u_{0,0}}$ & ${u_{0,1}}$ & ${u_{0,2}}$ & ${u_{0,3}}$ & 
$u_{0,0}$\\\hline\hline
& ${u_{0,0}}$ & $u_{0,1}$ & $u_{0,2}$ & $u_{0,3}$ &
\end{tabular}
\end{table}

From \eqref{eq:disc_inpaint} one can derive the linear systems of equations
\begin{equation} \label{eq:lsoe}
\underbrace{(\mathbf{C} - (\mathbf{I} - \mathbf{C}) \mathbf{A})}_{:= \mathbf{M}} 
\bm{u} = \mathbf{C} \bm{f}.
\end{equation}
This system of equations has a unique solution which fulfills a maximum-minimum 
principle. To prove this, we borrow the proof from \cite{Ma15}, chapter 2.4, 
which directly applies to our problem.
Its basic idea is to show positive definiteness of $\mathbf{M}$ using 
Gershgorin’s circle theorem \cite{Ge31} in combination with the block 
irreducibilty property discussed in \cite{FV62}.

%-------------------------------------------------------------------------------

\section{Application}
\label{sec:app}

Next, we want to combine antenna pointing information and microwave frequency 
feedback in order to visualize the reception quality of an S-band antenna for 
one year.
In our case, we focus on the signal information related to the reception of the 
DSCOVR satellite.
This comes down to a dataset containing $\sim$14.95 million measurements.
% 14 951 087 measurements, ~15 million measurements

We make use of a digital image plane -- which corresponds to the azimuth-elevation 
plane -- with a fixed resolution, e.g. 3600 $\times$ 901 px when using a 
precision of one for angles. At every position $(\varphi, \theta)_i$, we fill 
in the corresponding observed signal level.
Whenever overlaps of the input data -- due to resolution 
restrictions -- occur in the image plane, we average the input data.
Please note, that in all subsequent illustrations of signal level information, 
we apply an affine greyscale transformation for better visualization.
\begin{figure*}
\centering
\begin{tikzpicture}
\node [anchor=north west, inner sep=0pt] (img) at (current page.north west) {
\includegraphics[width=\textwidth]%
{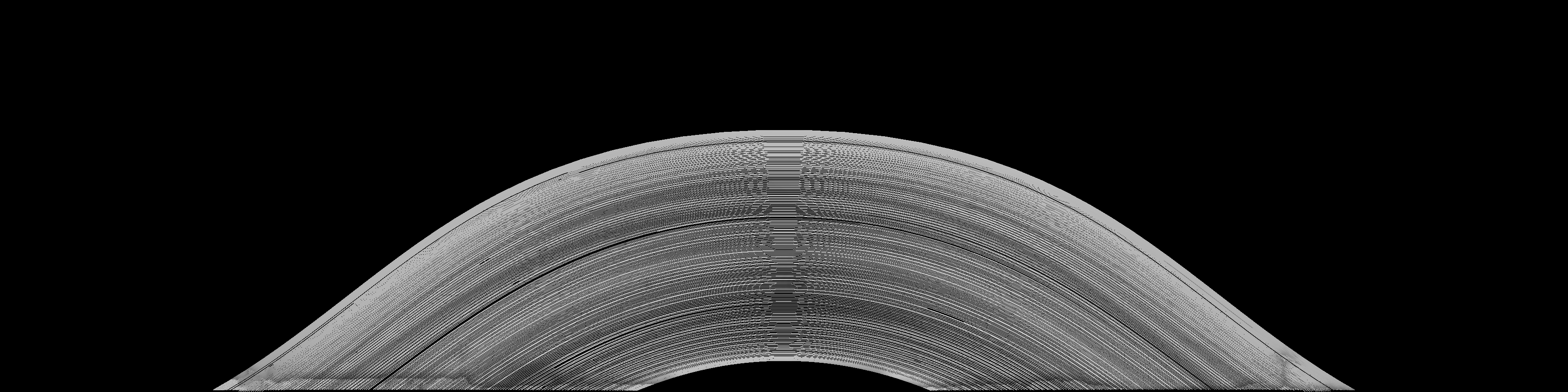}};
\node [%
minimum width=5mm,
minimum height=5mm,
inner sep=0pt,
anchor=west, 
text=white] 
(elevation) at (img.west) 
{$\theta$};
\node [%
minimum width=5mm,
minimum height=5mm,
inner sep=0pt,
anchor=north, 
text=white] 
(azimuth) at (img.north) 
{$\varphi$};
\node [%
minimum width=5mm,
minimum height=5mm,
inner sep=0pt,
anchor=north west, 
text=white]
(topleft) at (img.north west)
{};
\node [%
minimum width=5mm,
minimum height=5mm,
inner sep=0pt,
anchor=north east, 
text=white]
(topright) at (img.north east)
{};
\node [%
minimum width=5mm,
minimum height=5mm,
inner sep=0pt,
anchor=south west, 
text=white]
(bottomleft) at (img.south west)
{};
\node [anchor=north east, inner sep=0pt, yshift=-6mm] (zoom) at (img.north east) {
%trim={left bottom right top}
\includegraphics[width=.25\textwidth, trim={70cm 10cm 45cm 16cm}, clip]
{dscovr_quality_22121-23115_traj_mixed_bc_prec-1}};
%
% inner node
\node [anchor=east, inner sep=0pt, xshift=-20mm, yshift=-1mm, text=white] 
(X) at (zoom.south west)
{};
\draw[->, color=white] (elevation) -- (bottomleft);
\draw[->, color=white] (elevation) -- (topleft);
\draw[->, color=white] (azimuth) -- (topleft);
\draw[->, color=white] (azimuth) -- (topright);
%
% arrow
\draw[->, color=white] (X) -- (zoom.west);
\end{tikzpicture}
\vskip 1mm
\begin{tikzpicture}
\node [anchor=north west, inner sep=0pt] (img) at (current page.north west) {
\includegraphics[width=\textwidth]%
{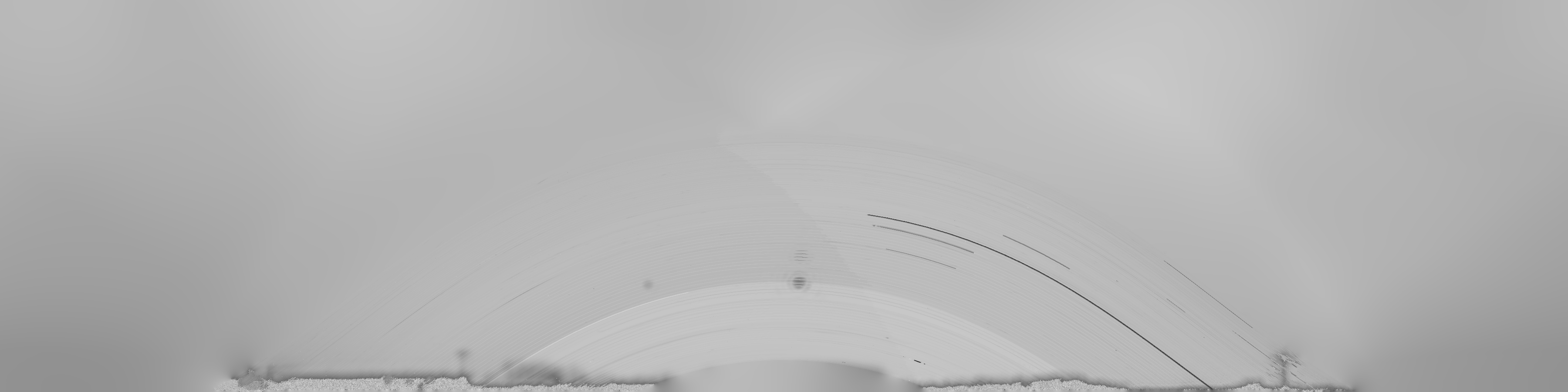}};
\node [%
minimum width=5mm,
minimum height=5mm,
inner sep=0pt,
anchor=north west, 
text=black]
(topleft) at (img.north west)
{};
\node [%
minimum width=5mm,
minimum height=5mm,
inner sep=0pt,
anchor=north east, 
text=black]
(topright) at (img.north east)
{};
\node [%
minimum width=5mm,
minimum height=5mm,
inner sep=0pt,
anchor=south west, 
text=black]
(bottomleft) at (img.south west)
{};
%
% node at X1
\node [anchor=south, inner sep=0pt, xshift=-35mm, yshift=2mm]
(X1) at (img.south)
{};
\node [%
minimum width=5mm,
minimum height=5mm,
inner sep=0pt,
anchor=west, 
text=black] 
(elevation) at (img.west) 
{$\theta$};
\node [%
minimum width=5mm,
minimum height=5mm,
inner sep=0pt,
anchor=north, 
text=black] 
(azimuth) at (img.north) 
{$\varphi$};
\draw[->, color=black] (elevation) -- (bottomleft);
\draw[->, color=black] (elevation) -- (topleft);
\draw[->, color=black] (azimuth) -- (topleft);
\draw[->, color=black] (azimuth) -- (topright);
\end{tikzpicture}
\caption{\textbf{Top:}~Image ($3600 \times 901$ pixels) showing trajectories of 
the DSCOVR satellite and the corresponding signal level information from the 
antenna's reception perspective. \textbf{Bottom:}~Result of 
homogeneous diffusion inpainting after 6412 iterations using $\varepsilon = 
10^{-8}$.}
\label{fig:rawtraj_and_inpainted}
\end{figure*}

In Fig.~\ref{fig:rawtraj_and_inpainted}, we show the signal level w.r.t. 
antenna 
pointing directions. The x-direction corresponds to the azimuth 
angles~$\varphi$, 
the y-direction to elevation angles~$\theta$ (with $\theta=0^\circ$ at the 
bottom and $\theta = 90^\circ$ at the top). Black areas indicate missing 
information.
On the top, one can see the satellite trajectories from the antenna's point of 
view combined with signal level information. The higher the signal level, 
the brighter the pixel.
There is one trajectory for every day of the year.
Since the DSCOVR satellite is positioned at the Sun-Earth $L_1$ Lagrange point, 
the trajecories roughly correspond to the position of the sun.
I.e. the trajectories with largest average elevation $\theta$ are close to 
summer solstice (azmiuth angles between 56 and 304 degrees). On the other 
hand, the lowest trajectories -- in terms of small average elevation $\theta$ 
-- are close to winter solstice (azmiuth angles between 139 and 221 degrees).

In the bottom image of Fig.\ref{fig:rawtraj_and_inpainted}, we illustrate the 
result of homogeneous diffusion inpainting. The inpainting keeps the known data 
untouched (inpainting mask set to one) and completes missing signal level 
information in the remaining areas.
Aside a trajectory with low signal level in the afternoon (the dark arc section 
on the 
right side of the illustration), the inpainted result emphasizes structures 
which 
can hardly be recognized in the raw data. This includes data at the bottom of the 
picture (which correspond to the ground), as well as features 
in the sky.

\begin{figure}
%trim={left bottom right top}
\includegraphics[width=\columnwidth, trim={20cm 0cm 85cm 26cm}, clip]
{dscovr_quality_22121-23115_img_mixed_bc_prec-1_00000001}\\[1mm]
\includegraphics[width=\columnwidth, trim={1.6cm 0.5cm 16.7cm 0.2cm}, clip]
{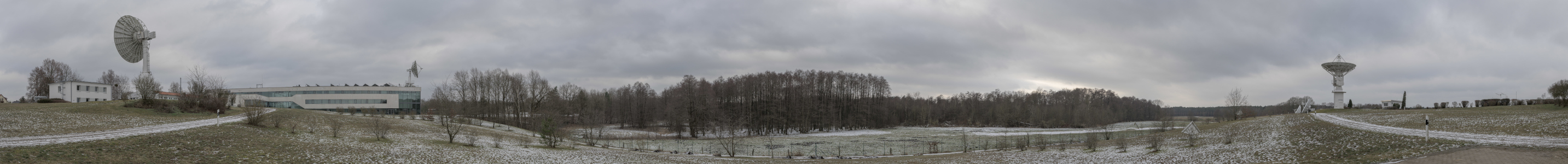}
\caption{\textbf{Top:}~Enlarged section of the inpainted signal level information 
for small elevation $\theta$ in the morning hours. 
\textbf{Bottom:}~Corresponding camera picture of the ground station 
from the antenna's point of view.}
\label{fig:gs_x1_x2}
\end{figure}

This becomes especially clear by taking a closer look at 
Fig.~\ref{fig:gs_x1_x2}. This excerpt of the inpainted signal level information 
from the morning hours and close to the ground contains significant shadows.
By taking a look at the corresponding camera picture, one can see that these 
shadows correspond to buildings and antenna systems of the ground station.
Even the trees on the right hand side are visible in the signal level information.

\begin{figure}
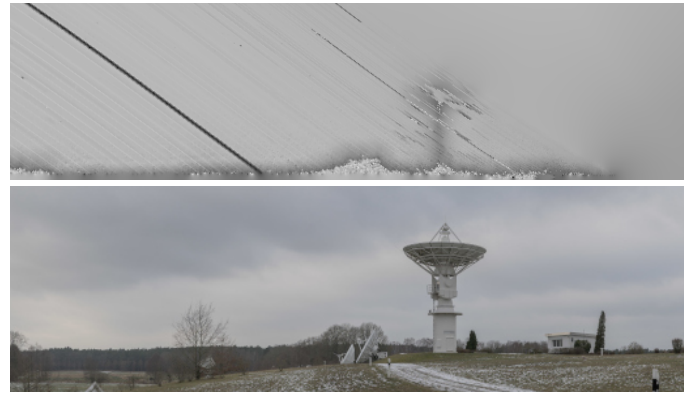

%trim={left bottom right top}
\includegraphics[width=\columnwidth, trim={90cm 0cm 15cm 26cm}, clip]
{dscovr_quality_22121-23115_img_mixed_bc_prec-1_00000001}\\[1mm]
\includegraphics[width=\columnwidth, trim={17.9cm 0.6cm 2cm 0.6cm}, clip]
{vt63_panorama-Jens_Richter-2024-01-20-small-300px}
\caption{\textbf{Top:}~Section of the inpainted signal level information for small 
elevation $\theta$ in the evening. \textbf{Bottom:}~Camera picture of 
the corresponding region from the antenna's point of view.}
\label{fig:tb1}
\end{figure}

Another example from the evening hours of satellite data reception 
close to the ground is given in Fig.~\ref{fig:tb1}.
Therein, we see that the signal quality gets degraded due to an antenna 
system impairing the signal line of sight.

\begin{figure}
    %trim={left bottom right top}
    \includegraphics[width=\columnwidth, trim={49cm 7cm 56cm 19cm}, clip]
    {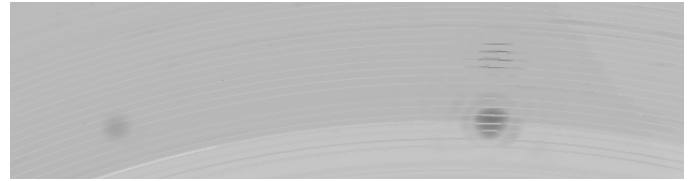}
    \caption{Highlighted drops in the signal level for an as yet unknown 
    reason.}
    \label{fig:drops}
\end{figure}

Apart from buildings, antenna systems, and natural obstacles, the inpainted signal 
information highlights further drops in the signal level around noon and in the 
late morning hours (see Fig.~\ref{fig:drops}, top).
Now, recall that this visualization offers a summary view over one year of data 
reception. This means that the degradation of the signal level occurred for 
almost identical antenna pointing angles for a number of subsequent days.
With the help of our visualization, we are also able to detect these 
-- in raw data -- easily overlooked events.
However, the reasons for this degradations are still under investigation by the 
ground station development team.

Apart from that, one can recognize a conspicuous pattern -- similar to a 
staircase -- at noon time throughout 
all trajectories where the signal level increases from left to right (see 
Fig.~\ref{fig:rawtraj_and_inpainted}). This effect correlates with a change 
in the satellite's data transmission rate which increases every day in this 
time period.

\subsection{Implementation Details and Runtime}
%
%while( snorm_rk > eps * snorm_r0 ) 
%As a stopping criterion use ||r^k||_2 / ||r^0||_2 <= epsilon
%
% Memory to be allocated for variable of type double... 8 Bytes.
%Memory to be allocated for system matrix M (naive implementation)... 78387.12 GiB 
%(80268408.00 MiB).
%Memory to be allocated for system matrix M... 0.12 GiB (123.73 MiB).
%Computation time... 61.524834452 seconds.
% Output image resolution... 3600 x 901 px
% parallelized version using 20 cores
%of a 12th Gen Intel(R) Core(TM) i7-12800H
%
We solve the linear system of equations \eqref{eq:lsoe} using the method of 
conjugate gradients \cite{HS52}.
As a stopping criterion, we employ a threshold of $\varepsilon$ on the relative 
residual decay in the $L^2$-norm.
We came up with an implementation in C which pays special attention to the 
sparseness of the matrix~$\mathbf{M}$. %$\mathbf{A}$. 
Consequently, the space and time complexity per iteration of the algorithm  is 
$\mathcal{O}(m)$, 
where $m$ denotes the number of nonzero entries of $\mathbf{M}$.
% discrete Laplacian, i.e. in our case 5 times N
The computation time for the result shown in Fig.~\ref{fig:rawtraj_and_inpainted}
was about 62~seconds using 20 threads of a Intel i7-12800H Processor.

%-------------------------------------------------------------------------------

\section{Summary and Outlook}
\label{sec:concl}
Our paper shows the high potential of inpainting techniques for technical signal 
information, especially, in the context of satellite data reception.
The visualization of the gained quality information product allows the 
identification and treatment of potential problems within our satellite data 
reception chain which might be overseen otherwise.

In our future work, we will take a look at the solution of the inpainting 
problem 
in different coordinate systems. In terms of antenna pointing directions, 
spherical coordinates might be a more appropriate approach.
This, however, comes 
at the cost of a more comprehensive well-posedness analysis regarding the 
underlying mathematical problem.

Furthermore,
we also plan to expand our inpainting approach to support input data 
from additional sources like digital images. A sensor fusion strategy is highly 
likely to cope better with missing information.
Connected to this idea is the consideration of anisotropic inpainting methods.
In this way, existing structures within the input data could be better 
reflected in the result.

Our contribution has already proven to be a valuable tool for the evaluation of 
satellite data reception and the development of the ground station Neustrelitz.

%-------------------------------------------------------------------------------

\section*{Acknowledgment}
I thank my colleagues at DLR, especially Robert Miesen and Jens Richter 
for valuable comments and fruitful discussions. The latter has also provided 
the photos being used in this paper.

%-------------------------------------------------------------------------------

\bibliographystyle{IEEEtran}
\bibliography{eusipco_2024.bib}

\end{document}